\documentclass[12pt]{article}

\usepackage{amsmath}
\usepackage{amssymb}
\usepackage{booktabs}
\usepackage{graphicx}
\usepackage{hyperref}
\usepackage[numbers]{natbib}
\usepackage[a4paper, margin=2.5cm]{geometry}
\usepackage{xcolor}
\usepackage{array}
\usepackage{tabularx}
\usepackage{float}
\usepackage{subcaption}
\usepackage{multirow}

\hypersetup{
    colorlinks=true,
    linkcolor=blue,
    citecolor=blue,
    urlcolor=blue
}

\title{Enhanced Mycelium of Thought (EMoT):\\ A Bio-Inspired Hierarchical Reasoning Architecture with Strategic Dormancy and Mnemonic Encoding}

\author{
    Florian Odi Stummer$^{1}$ \\[6pt]
    $^1$ Institute of General Medicine\\
    Martin Luther University Halle-Wittenberg, Germany \\    
    \textbf{Correspondence:} florian.stummer@uk-halle.de
}

\date{}

\begin{document}

\maketitle

% ============================================================
% ABSTRACT
% ============================================================
\begin{abstract}
Current prompting paradigms for large language models (LLMs), including Chain-of-Thought (CoT) and Tree-of-Thoughts (ToT), follow linear or tree-structured reasoning paths that lack persistent memory, strategic dormancy, and cross-domain synthesis capabilities. We present the Enhanced Mycelium of Thought (EMoT) framework, a bio-inspired reasoning architecture that organises cognitive processing into a four-level hierarchy (Micro, Meso, Macro, Meta), implements strategic dormancy and reactivation of reasoning nodes, and integrates a Memory Palace with five mnemonic encoding styles for persistent knowledge retention. EMoT is positioned as a research prototype for complex, multi-domain problems, not as a general-purpose prompting enhancement. The architecture is realised in a 3,150-line Python implementation with a pluggable LLM backend and 51 regression tests. Two complementary evaluations reveal a characteristic trade-off. In a blind LLM-as-Judge evaluation across three domains with method labels removed, EMoT achieved near-parity with CoT (4.20 vs.\ 4.33/5.0) with higher run-to-run stability, and outperformed CoT on Cross-Domain Synthesis (4.8 vs.\ 4.4). Ablation studies demonstrate that strategic dormancy is architecturally essential (quality collapsed from 4.2 to 1.0 when disabled). On a complementary 15-item short-answer benchmark, EMoT (27\% accuracy) substantially underperformed simpler baselines, confirming systematic overthinking on simple problems. These preliminary results are subject to important limitations: small sample sizes ($n=3$ complex cases, $n=15$ short-answer items), reliance on LLM-as-Judge evaluation with potential self-preference bias, and approximately 33-fold computational cost overhead relative to CoT. To our knowledge, EMoT is the first reasoning framework to combine hierarchical network topology, strategic thought dormancy with reactivation, and mnemonic memory encoding in an integrated architecture.

\end{abstract}

% ============================================================
% 1. INTRODUCTION
% ============================================================
\section{Introduction}

The capacity of large language models to perform complex reasoning has improved substantially through structured prompting paradigms. Chain-of-Thought (CoT) prompting \cite{ref1} elicits step-by-step reasoning, while Tree-of-Thoughts (ToT) \cite{ref2} enables branching exploration with search-based path selection. Graph-of-Thoughts (GoT) \cite{ref3} extends this further by permitting arbitrary graph topologies. Self-Consistency \cite{ref4} improves robustness through majority voting over multiple reasoning paths, and Reflexion \cite{ref5} introduces verbal reinforcement learning for iterative self-improvement.

Despite these advances, current frameworks share several fundamental limitations. First, they are structurally constrained: CoT is strictly linear, ToT is hierarchically branching, and even GoT, while permitting cycles, does not maintain persistent state across reasoning episodes. Second, existing paradigms employ static pruning: ToT permanently discards reasoning paths deemed unpromising, preventing the reconsideration of initially undervalued hypotheses when new evidence emerges. Third, no current framework provides a persistent memory mechanism that encodes, stores, and retrieves insights across reasoning iterations using cognitively grounded encoding strategies.

These limitations are particularly consequential in domains characterised by diagnostic uncertainty, multi-system interactions, and evolving evidence. In clinical medicine, for instance, a symptom initially attributed to one pathology may, upon further investigation, point to an entirely different mechanism. A reasoning system that irrevocably prunes such alternative hypotheses loses the capacity for the kind of diagnostic reconsideration that experienced clinicians routinely perform \cite{ref13}. Croskerry's dual-process model of clinical reasoning \cite{ref13} distinguishes between rapid, heuristic-driven Type~1 processing and deliberate, analytical Type~2 processing; diagnostic errors frequently occur at the transition between these modes, particularly when Type~1 processing prematurely closes the diagnostic search. EMoT's dormancy mechanism provides a computational analogue to this cognitive transition: rather than allowing initial heuristic judgements to permanently close reasoning paths, dormant nodes preserve alternative hypotheses that can be reactivated when deliberate analysis reveals inconsistencies with the initial assessment. Croskerry \cite{ref37} further argued for ``diagnostic timeouts'' as a deliberate metacognitive strategy to interrupt premature closure; EMoT's dormancy-reactivation cycle operationalises this concept computationally.

Biological systems offer a compelling alternative model. Mycelial networks, the underground filamentous structures of fungi, exhibit properties consistent with decentralised information processing and adaptive resource allocation \cite{ref6, ref7, ref28}. They dynamically allocate resources across vast spatial networks, maintain redundant pathways that can be reactivated when conditions change, facilitate inter-organism communication, and solve complex optimisation problems without centralised control \cite{ref8, ref9}. Simard et al.\ \cite{ref16} further documented how mycorrhizal networks mediate inter-plant communication and resource sharing across forest ecosystems. Recent work has demonstrated that mycelium networks are capable of implementing logical gates through propagation of electrical signals \cite{ref21}, and mycelium-inspired neural network architectures have shown promising feature extraction capabilities in geoscience data mining tasks \cite{ref22}, validating the computational relevance of this biological metaphor. These properties, including adaptive resource allocation, dormancy with context-sensitive reactivation, and cross-network synthesis, map directly onto desirable characteristics for reasoning architectures.

In this paper, we present the Enhanced Mycelium of Thought (EMoT) framework, a reasoning architecture inspired by the structural and functional properties of mycelial networks. We position EMoT explicitly as a research prototype of reasoning infrastructure for complex, multi-domain problems, not as a general-purpose prompting enhancement or a competitor to established accuracy benchmarks. EMoT makes three principal contributions:

\begin{enumerate}
    \item \textbf{A hierarchical network architecture} with four processing levels (Micro, Meso, Macro, Meta) that supports both bottom-up insight generation and top-down constraint propagation, augmented by five specialised enhancement modules.

    \item \textbf{Strategic dormancy and reactivation mechanisms} that preserve lower-confidence reasoning nodes in a dormant state rather than pruning them, enabling context-sensitive reactivation when new evidence alters the relevance landscape.

    \item \textbf{An integrated Memory Palace} with five mnemonic encoding styles (Visual Hook, Loci Room, Chunking, Temporal Ladder, Narrative Hook) that provides persistent, retrievable memory across reasoning iterations.
\end{enumerate}

We provide a complete open-source implementation (3,150 lines of Python, 51 regression tests) with a pluggable LLM backend, and present preliminary empirical results from benchmark evaluations comparing EMoT to standard CoT prompting across three problem domains.

% ============================================================
% 2. RELATED WORK
% ============================================================
\section{Related Work}

\subsection{Prompting-Based Reasoning Frameworks}

\textbf{Chain-of-Thought (CoT).} Wei et al.\ \cite{ref1} demonstrated that prompting LLMs to produce intermediate reasoning steps substantially improves performance on arithmetic, commonsense, and symbolic reasoning tasks. Zero-shot CoT \cite{ref10} showed that simply appending ``Let's think step by step'' activates latent reasoning capabilities. However, CoT is inherently sequential: it follows a single reasoning path without mechanisms for backtracking, parallel exploration, or reconsideration of earlier steps.

\textbf{Tree-of-Thoughts (ToT).} Yao et al.\ \cite{ref2} extended CoT by framing reasoning as a search problem over a tree of intermediate ``thoughts,'' using breadth-first or depth-first search with LLM-based evaluation to select promising branches. While ToT enables exploration of multiple solution paths, it employs static pruning that permanently eliminates branches scoring below a threshold, precluding their later reconsideration. Furthermore, ToT's tree structure prevents cross-branch connections that might yield novel synthetic insights.

\textbf{Graph-of-Thoughts (GoT).} Besta et al.\ \cite{ref3} generalised the topology further, modelling reasoning as an arbitrary directed graph where thoughts can be combined, refined, or looped. GoT represents a significant structural advance but does not address persistent memory or dormancy.

\textbf{Self-Consistency.} Wang et al.\ \cite{ref4} proposed sampling multiple diverse reasoning paths and selecting the most consistent answer through majority voting. This approach improves robustness but does not preserve or reactivate individual reasoning traces.

\textbf{Reflexion.} Shinn et al.\ \cite{ref5} introduced a framework where language agents maintain a verbal memory of previous attempts and their outcomes, using this reflective feedback to improve subsequent performance. Reflexion shares EMoT's emphasis on learning from past reasoning but operates at the level of complete task attempts rather than at the granularity of individual reasoning nodes.

\textbf{Recent advances (2025).} The reasoning framework landscape has continued to evolve rapidly. Pandey et al.\ \cite{ref17} introduced Adaptive Graph of Thoughts (AGoT), which unifies chain, tree, and graph paradigms through dynamic DAG decomposition, achieving up to 46.2\% improvement on scientific reasoning tasks. Hao et al.\ \cite{ref18} proposed RL of Thoughts (RLoT), using a lightweight reinforcement learning navigator to dynamically select logic blocks, outperforming established inference-time techniques by up to 13.4\%. Tang et al.\ \cite{ref19} presented Matrix of Thought, exploring horizontal and vertical reasoning dimensions through column-cell communication. While these frameworks advance the state of the art in adaptive reasoning structure, none incorporates persistent memory across reasoning iterations, strategic dormancy with reactivation, or mnemonic encoding of intermediate insights; these remain the distinguishing contributions of EMoT.

\textbf{Reasoning-optimised LLMs.} A parallel development has been the emergence of models with internalised reasoning capabilities. OpenAI's o1 and o3 models \cite{ref35} employ extended chain-of-thought during inference, producing substantially improved performance on mathematical and scientific reasoning benchmarks. DeepSeek-R1 \cite{ref36} demonstrated that reinforcement learning can elicit sophisticated reasoning behaviours including self-verification and backtracking, achieving performance competitive with o1 on several benchmarks. These approaches differ fundamentally from EMoT in that they internalise reasoning strategies within model weights and hidden chain-of-thought, whereas EMoT operates as an external scaffold that structures reasoning across multiple LLM calls with explicit, inspectable state management. The two approaches are complementary: EMoT could use a reasoning-optimised model as its LLM backend, potentially combining internalised reasoning depth with EMoT's persistent memory and dormancy management. We note, however, that the opacity of internalised reasoning chains contrasts with EMoT's architectural transparency, where every node, dormancy transition, and memory encoding is explicitly logged and auditable.

\textbf{Axis of novelty.} To clarify EMoT's positioning relative to these recent advances: AGoT \cite{ref17} and RLoT \cite{ref18} achieve test-time structural adaptation by dynamically selecting reasoning topologies, but neither maintains persistent memory across reasoning iterations nor implements dormancy with context-sensitive reactivation. Reasoning-optimised LLMs such as o1 \cite{ref35} and DeepSeek-R1 \cite{ref36} internalise extended reasoning within model weights and hidden chain-of-thought, whereas EMoT operates as an external scaffold with explicit, inspectable state. EMoT's specific axis of novelty lies in the combination of three architectural primitives: strategic dormancy with reactivation, mnemonic encoding for persistent memory, and hierarchical multi-level processing; to our knowledge, no existing framework integrates all three.

\subsection{Cognitive Architectures}

Classical cognitive architectures such as SOAR \cite{ref11} and ACT-R \cite{ref12} provide formal models of human cognition with explicit memory systems, goal management, and learning mechanisms. SOAR's universal subgoaling and chunking mechanisms share conceptual similarities with EMoT's hierarchical decomposition and Memory Palace, respectively. ACT-R's distinction between declarative and procedural memory aligns with EMoT's separation of insight storage (Memory Palace) from processing logic (enhancement modules). However, these architectures were designed for symbolic AI systems rather than as reasoning scaffolds for LLMs.

\subsection{Biological Models of Distributed Intelligence}

The biological inspiration for EMoT draws from research on mycelial networks and fungal intelligence. Sheldrake \cite{ref6} documents how mycorrhizal networks connect multiple plants, facilitating resource sharing and inter-organism signalling. Fricker et al.\ \cite{ref7} formalised the adaptive network properties of fungal mycelia, showing how they solve transport optimisation problems through decentralised mechanisms. Gorzelak et al.\ \cite{ref8} demonstrated that mycorrhizal networks mediate complex adaptive behaviour in plant communities. Adamatzky \cite{ref9} reported on the computational properties of fungal networks, including their capacity for spiking-like information transmission. Tero et al.\ \cite{ref28} showed that the slime mould \textit{Physarum polycephalum} constructs transport networks comparable in efficiency to human-designed infrastructure, providing experimental evidence for biologically inspired adaptive network design.

These biological systems exhibit properties that are absent from current reasoning frameworks: (a) persistent network structure with dormant pathways that can be reactivated; (b) multi-scale organisation from local to global; (c) cross-network resource and information transfer; and (d) adaptive reorganisation in response to environmental change. EMoT translates these biological properties into computational mechanisms for LLM-based reasoning.

\subsection{Clinical Decision Support Context}

EMoT is a reasoning architecture, not a clinical decision support system (CDSS). Established CDSS platforms such as DXplain \cite{ref29} and Isabel Healthcare \cite{ref30} provide diagnosis support through curated medical knowledge bases and have undergone extensive clinical validation. More recently, large language model-based systems such as Med-PaLM~2 \cite{ref31} have demonstrated strong performance on medical question-answering benchmarks, though their deployment in clinical settings remains limited. EMoT differs from these systems in that it provides a general-purpose reasoning architecture that could, in principle, underpin a CDSS, but it has not itself undergone the clinical validation, safety testing, or regulatory evaluation required for clinical deployment. The clinical case presented in this paper serves as an illustrative demonstration of cross-domain reasoning, not as evidence of clinical utility.

\subsection{How EMoT Differs}

Table~\ref{tab:framework_comparison} summarises the key architectural differences between EMoT and existing frameworks.

\begin{table}[!ht]
\centering
\caption{Comparison of reasoning framework properties.}
\label{tab:framework_comparison}
\small
\begin{tabularx}{\columnwidth}{@{}lXXXXX@{}}
\toprule
\textbf{Property} & \textbf{CoT} & \textbf{ToT} & \textbf{GoT} & \textbf{Reflexion} & \textbf{EMoT} \\
\midrule
Topology & Linear & Tree & Graph & Linear + Memory & Hierarch.\ network \\
Parallel paths & No & Yes & Yes & No & Yes \\
Thought preserv. & No & No (pruned) & No & Verbal summary & Dormant nodes \\
Reactivation & No & No & No & Partial & Yes (ctx-sens.) \\
Persistent mem. & No & No & No & Verbal buffer & Mem.\ Palace (5 enc.) \\
Cross-dom.\ synth. & No & Limited & Possible & No & Explicit (IDE) \\
Eval.\ dimensions & None & 1 (LLM) & 1 & Binary (s/f) & 4 (S,N,D,C) \\
Hierarch.\ levels & 1 & Variable & Variable & 1 & 4 (fixed) \\
\bottomrule
\end{tabularx}
\normalsize
\end{table}

% ============================================================
% 3. THE EMoT FRAMEWORK
% ============================================================
\section{The EMoT Framework}

\subsection{Hierarchical Network Architecture}

EMoT organises reasoning into a four-level hierarchical network, where each level serves a distinct cognitive function:

\begin{itemize}
    \item \textbf{Micro Level (Detail):} Processes specific facts, data points, and atomic observations. Micro nodes generate basic insights from individual pieces of evidence and propagate them upward. In the clinical reasoning context, a Micro node might process a single laboratory value (e.g., ``B12 = 291 pmol/L, within the clinically ambiguous zone of 150--300 pmol/L'' \cite{ref32}).

    \item \textbf{Meso Level (Pattern):} Identifies relationships and patterns across multiple Micro-level insights. Meso nodes perform correlation detection, temporal sequencing, and causal hypothesis generation. A Meso node might connect the B12 level with the history of supplement discontinuation and the onset of neurological symptoms.

    \item \textbf{Macro Level (Solution):} Synthesises Meso-level patterns into coherent solution components. Macro nodes integrate evidence across domains to produce diagnostic hypotheses, policy proposals, or strategic recommendations.

    \item \textbf{Meta Level (Strategy):} Provides oversight, quality assurance, and strategic guidance. Meta nodes evaluate the overall coherence of the reasoning process, identify gaps, and direct resource allocation to underexplored areas.
\end{itemize}

Information flows both bottom-up (insight aggregation from Micro to Meta) and top-down (constraint propagation from Meta to Micro). Lateral connections within each level enable cross-domain pattern matching. This bidirectional, multi-scale processing distinguishes EMoT from the strictly bottom-up aggregation of ToT and the flat topology of CoT. EMoT's triangulation arises from combining hierarchical depth, cross-level feedback, and persistent memory, whereas ToT's triangulation is limited to search over branches with static pruning and no mechanism for reactivating previously discarded paths. Figure~\ref{fig:reasoning_trace} illustrates a representative reasoning trace for the Patient Bengt case, showing how dormant nodes are preserved across iterations and reactivated when new evidence warrants reconsideration.

\subsection{Node Processing with LLM Integration}

Each node in the EMoT network encapsulates a reasoning unit that can optionally invoke an LLM for content generation. The system implements a pluggable \texttt{LLMBackend} abstraction supporting four providers:

\begin{itemize}
    \item \textbf{Anthropic} (Claude API): Used for high-quality reasoning in production benchmarks.
    \item \textbf{Google} (Gemini): Alternative cloud provider.
    \item \textbf{Ollama} (local models, e.g., Qwen-3 14B): For on-premise deployment where data privacy is paramount.
    \item \textbf{Stub} (template-based): For testing and development without API costs.
\end{itemize}

Node processing follows a three-phase cycle: (1)~\textbf{Perception}, where the node receives inputs from connected nodes and the current problem context; (2)~\textbf{Processing}, where the node generates insights through LLM invocation or rule-based logic; and (3)~\textbf{Propagation}, where outputs are transmitted to connected nodes at the same or adjacent hierarchical levels.

The \texttt{LLMBackend} class tracks call counts and token usage, enabling the Computational Efficiency Optimiser (CEO) module to monitor and manage resource consumption across the network.

\subsection{Strategic Dormancy Controller (SDC)}

The Strategic Dormancy Controller is arguably EMoT's most distinctive feature. Unlike ToT's static pruning, which permanently eliminates low-scoring branches, EMoT's SDC implements a three-mechanism approach to managing underperforming reasoning nodes:

\begin{enumerate}
    \item \textbf{Predictive Relevance Modelling:} When a node's Trust Score falls below a threshold (default: $T < 0.5$), it is not deleted but transitioned to a dormant state. The SDC maintains metadata about the node's content, the context in which it was created, and a predicted relevance profile that estimates conditions under which reactivation might be warranted.

    \item \textbf{Partial Activation Mechanism:} Dormant nodes are not fully inactive. The SDC periodically evaluates dormant nodes against the current reasoning context, computing a reactivation probability. Nodes can be partially activated, contributing their insights at reduced weight, before full reactivation is committed.

    \item \textbf{Temporal Reasoning Engine:} The SDC tracks the temporal evolution of the reasoning process, identifying phase transitions (e.g., from hypothesis generation to hypothesis testing) that may alter the relevance of dormant nodes. A differential diagnosis initially dismissed in favour of a more probable explanation might be reactivated when first-line treatment fails to produce expected improvement.
\end{enumerate}

\subsection{Memory Palace and Mnemonic Encoding}

The Memory Palace is EMoT's persistent storage mechanism, inspired by the method of loci from classical mnemonic techniques \cite{ref14}. It stores insights, intermediate results, and cross-domain connections in five encoding styles:

\begin{enumerate}
    \item \textbf{Visual Hook:} Encodes insights as vivid, image-like descriptions that facilitate pattern matching through visual similarity.
    \item \textbf{Loci Room:} Organises insights spatially within a virtual ``room,'' where position encodes relational structure.
    \item \textbf{Chunking:} Groups related insights into compressed, retrievable units that reduce cognitive load.
    \item \textbf{Temporal Ladder:} Arranges insights chronologically, preserving the temporal sequence of discovery and enabling temporal reasoning.
    \item \textbf{Narrative Hook:} Encodes insights within a narrative structure that preserves causal relationships and contextual meaning.
\end{enumerate}

The neuroscientific basis for this approach is supported by neuroimaging evidence. Dresler et al.\ \cite{ref23} demonstrated through fMRI that mnemonic training reshapes brain networks to support superior memory, with memory athletes showing distinct functional connectivity patterns compared to untrained controls. Ren et al.\ \cite{ref33} further showed that method of loci training yields unique prefrontal representations that support effective memory encoding. Blunt and VanArsdall \cite{ref24} showed additive effects of the method of loci with animacy-based encoding ($N=354$), confirming that structured mnemonic strategies produce measurably higher recall rates.

The diversity of encoding styles serves a functional purpose: different retrieval contexts may benefit from different encoding formats. A clinical reasoning task might benefit from Temporal Ladder encoding (when did symptoms first appear?) and Chunking (grouping related laboratory values), while a policy design task might benefit from Narrative Hook encoding (the causal chain from problem to proposed intervention).

\subsection{Enhancement Modules}

Five specialised modules address specific reasoning challenges:

\textbf{Quality Amplification Module (QAM):} Implements a progressive refinement pipeline that iteratively improves solution quality through multi-stage processing. QAM incorporates feedback from higher-level nodes, enforces consistency with problem constraints, and calibrates confidence levels based on evidence strength.

\textbf{Insight Distillation Engine (IDE):} Transforms the high volume of insights generated by the network into high-quality, actionable knowledge. IDE performs insight clustering (using DBSCAN for semantic grouping via TF-IDF vectorisation), relevance scoring, contradiction resolution, and utilisation tracking.

\textbf{Computational Efficiency Optimiser (CEO):} Manages resource allocation across the network, implementing pathway pruning for low-value reasoning paths, result caching for node outputs, and parallel processing optimisation for independent reasoning branches.

\textbf{Hierarchical Integration Framework (HIF):} Coordinates information flow across hierarchical levels through multi-scale representation, bottom-up aggregation, top-down constraint propagation, and lateral integration within levels.

\textbf{Strategic Dormancy Controller (SDC):} As described in Section~3.3, manages the lifecycle of dormant nodes through predictive relevance modelling, partial activation, and temporal reasoning.

\subsection{Trust Score}

Each reasoning node is evaluated across four dimensions, combined into a composite Trust Score:

\begin{equation}
T = 0.4 \cdot S + 0.2 \cdot N + 0.2 \cdot D + 0.2 \cdot C
\end{equation}

\noindent where:
\begin{itemize}
    \item \textbf{S} (Success Likelihood): Estimated probability of the node contributing to a correct or useful solution (range 0--1).
    \item \textbf{N} (Novelty): Degree to which the insight represents an original or non-obvious contribution (range 0--1).
    \item \textbf{D} (Depth): Comprehensiveness and analytical depth of the reasoning (range 0--1).
    \item \textbf{C} (Coherence): Internal logical consistency and integration with other nodes (range 0--1).
\end{itemize}

The asymmetric weighting reflects a design priority: Success Likelihood receives double the weight of each other dimension, embodying the principle that reasoning quality is primarily judged by its contribution to problem resolution, with novelty, depth, and coherence serving as secondary quality indicators. This weighting scheme is heuristic in nature, chosen during development based on the authors' judgement that solution correctness should dominate the composite score. Alternative weightings (e.g., equal weights, or increased weight on coherence for tasks requiring internal consistency) may be preferable in different application contexts. The Trust Score threshold for dormancy ($T < 0.5$) was similarly set heuristically during development and retained after empirical testing confirmed reasonable dormancy behaviour across benchmark cases. The weights and threshold ($T < 0.5$) are heuristic; no systematic hyperparameter search or sensitivity analysis was conducted. Exploring these parameters systematically remains an important direction for future work.

% ============================================================
% 4. IMPLEMENTATION
% ============================================================
\section{Implementation}

\subsection{System Architecture}

The EMoT framework is implemented as a monolithic Python module (\texttt{MoT\_5\_0.py}, 3,150 lines of code) with supporting modules for advanced scoring (40 LOC), stability enhancement (38 LOC), and a reasoning engine abstraction layer (186 LOC). The implementation uses NumPy for numerical operations, NetworkX for graph representation, and scikit-learn (DBSCAN, TF-IDF) for insight clustering within the IDE module. All benchmarks reported in this paper were run with Python 3.14, NumPy 2.4, NetworkX 3.6, and scikit-learn 1.8.

The total codebase, including benchmarking infrastructure and tests, comprises approximately 6,700 lines of Python across 10 source files. The benchmark suite (\texttt{bench/}) includes three evaluation scripts: a quality benchmark comparing EMoT to CoT (668 LOC), a cross-model benchmark (876 LOC), and a multi-technique reasoning benchmark (1,151 LOC).

\subsection{Test Suite}

The implementation is validated by 51 regression tests covering core functionality including node creation, hierarchical processing, dormancy transitions, Memory Palace encoding and retrieval, Trust Score computation, LLM backend abstraction, and enhancement module behaviour. All 51 tests pass on the current codebase.

\subsection{Pluggable LLM Backend}

The \texttt{LLMBackend} class implements a singleton-pattern abstraction over four LLM providers, configured via environment variables (\texttt{MOT\_LLM\_BACKEND}, \texttt{MOT\_LLM\_MODEL}) or constructor parameters. Default models are Claude Sonnet~4 (Anthropic), Gemini 2.0 Flash (Google), Qwen-3 14B (Ollama), and a deterministic stub for testing. The backend tracks cumulative call counts and token usage, enabling cost estimation and efficiency analysis.

% ============================================================
% 5. EVALUATION
% ============================================================
\section{Evaluation}

\subsection{Evaluation Framework}

We evaluate EMoT using an LLM-as-Judge methodology, where an independent LLM instance (Claude Sonnet~4) rates the quality of reasoning outputs on a 6-criterion rubric, each scored on a 1--5 scale:

\begin{enumerate}
    \item \textbf{Recursion Depth:} Degree of hierarchical problem decomposition.
    \item \textbf{Dormant Thought Management:} Evidence of strategic hypothesis preservation and reactivation.
    \item \textbf{Cross-Domain Synthesis:} Integration of insights across multiple knowledge domains.
    \item \textbf{Memory Utilisation:} Coherent reference to and building upon earlier reasoning.
    \item \textbf{Structured Output:} Organisation, formatting, and readability.
    \item \textbf{Solution Quality:} Completeness, correctness, and actionability of the proposed solution.
\end{enumerate}

The LLM-as-Judge paradigm is increasingly used for evaluating open-ended generative tasks where ground-truth answers are not available \cite{ref15}. We acknowledge that this approach introduces potential biases, including the model's preference for its own output style, which we address in the limitations section. We also note that two of these criteria -- Dormant Thought Management and Memory Utilisation -- reflect architecture-specific mechanisms and may inherently favour EMoT relative to frameworks that do not explicitly model dormancy or persistent memory. Future evaluations should consider architecture-neutral rubrics to enable fairer comparisons.

\textbf{Threats to validity.} Several limitations of the LLM-as-Judge setup should be noted upfront. First, the judge model (Claude Sonnet~4) belongs to the same model family as the generator, introducing a potential self-preference bias where outputs with lower perplexity receive inflated scores regardless of origin \cite{ref20}. Second, the inclusion of Dormant Thought Management and Memory Utilisation in the rubric creates a risk of circularity, as these criteria describe EMoT-specific mechanisms (see Section~6.5 for extended discussion). Third, the sample size is small ($n=3$ cases, 3 runs per condition), limiting statistical power. Fourth, using a single judge model precludes cross-model agreement checks that would strengthen score reliability.

For each evaluation, both EMoT and the baseline CoT approach receive identical problem prompts. EMoT processes the problem through its full hierarchical architecture with three iterations, while CoT receives a single-pass prompt instructing step-by-step reasoning. The judge model scores both outputs blind to their source: method labels are removed from outputs before evaluation, and the judge prompt does not identify which output came from which system. Each condition is evaluated in three independent runs to provide variance estimates.

Given the small number of test cases ($n=3$), these results should be treated as descriptive observations from a preliminary, exploratory study. The three runs per condition provide variance estimates but do not support formal statistical inference.

\subsection{Quality Benchmark: Blind-Judge Evaluation}

We evaluated EMoT against standard CoT on three test cases spanning distinct domains: (1) clinical reasoning (a geriatric multi-morbidity case), (2) policy design (global climate migration strategy), and (3) AI governance (pandemic vaccine prioritisation). All evaluations used Claude Sonnet~4 as the reasoning model and an independent Anthropic model instance as the judge. To mitigate evaluation bias, method labels were removed from outputs before judging: the judge evaluated two anonymised responses per case without knowledge of which system produced which output. Each condition was evaluated in three independent runs to provide variance estimates. Given the small sample size ($n=3$ cases), these results should be considered exploratory and hypothesis-generating rather than definitive.

\begin{table}[!ht]
\centering
\caption{Per-run overall scores: blind-judge quality benchmark.}
\label{tab:per_run_scores}
\begin{tabular}{lccc}
\toprule
\textbf{Run} & \textbf{EMoT Score} & \textbf{CoT Score} & \textbf{Winner} \\
\midrule
Run 1 & 4.2 & 4.2 & Tie \\
Run 2 & 4.2 & 4.5 & CoT \\
Run 3 & 4.2 & 4.3 & CoT \\
\midrule
\textbf{Mean (SD)} & \textbf{4.20 (0.00)} & \textbf{4.33 (0.15)} & \textbf{CoT} \\
\bottomrule
\end{tabular}
\end{table}

\begin{table}[!ht]
\centering
\caption{Per-criterion means across three blind evaluation runs (mean scores, 1--5 scale).}
\label{tab:per_criterion_means}
\begin{tabular}{lccc}
\toprule
\textbf{Criterion} & \textbf{EMoT Mean} & \textbf{CoT Mean} & \textbf{Winner} \\
\midrule
Recursion Depth & 4.0 & 4.6 & CoT \\
Dormant Thought Management & 2.9 & 3.1 & CoT \\
Cross-Domain Synthesis & 4.8 & 4.4 & EMoT \\
Memory Utilisation & 4.0 & 4.2 & CoT \\
Structured Output & 5.0 & 5.0 & Tie \\
Solution Quality & 4.6 & 4.7 & CoT \\
\midrule
\textbf{Overall} & \textbf{4.20} & \textbf{4.33} & \textbf{CoT} \\
\bottomrule
\end{tabular}
\end{table}

\begin{figure}[!ht]
\centering
\includegraphics[width=\columnwidth]{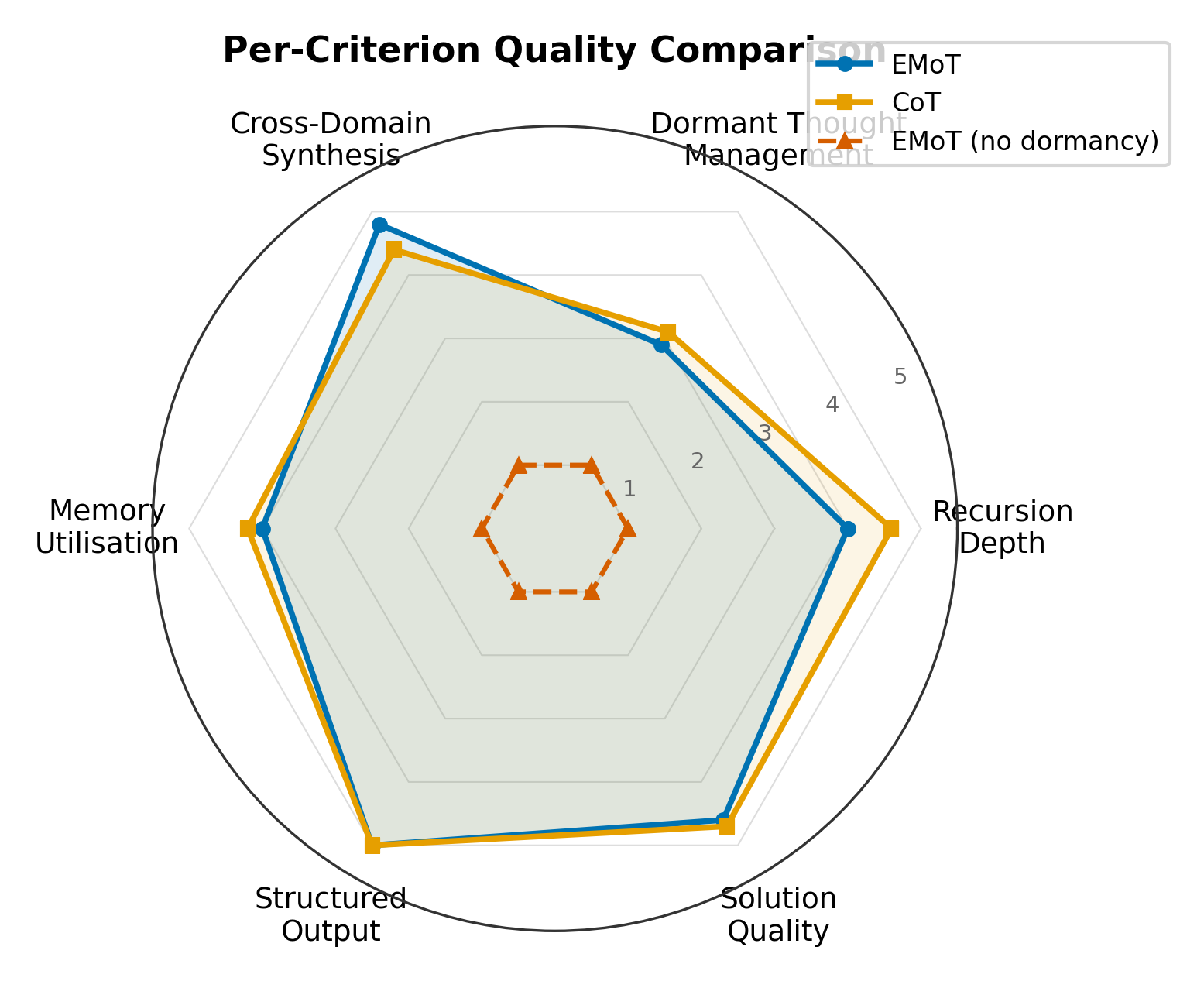}
\caption{Radar chart comparing EMoT and CoT across six evaluation criteria. EMoT outperforms CoT on Cross-Domain Synthesis (4.8 vs.\ 4.4) while CoT leads on Recursion Depth and Solution Quality. Both achieve parity on Structured Output.}
\label{fig:radar}
\end{figure}

\begin{figure}[!ht]
\centering
\includegraphics[width=\columnwidth]{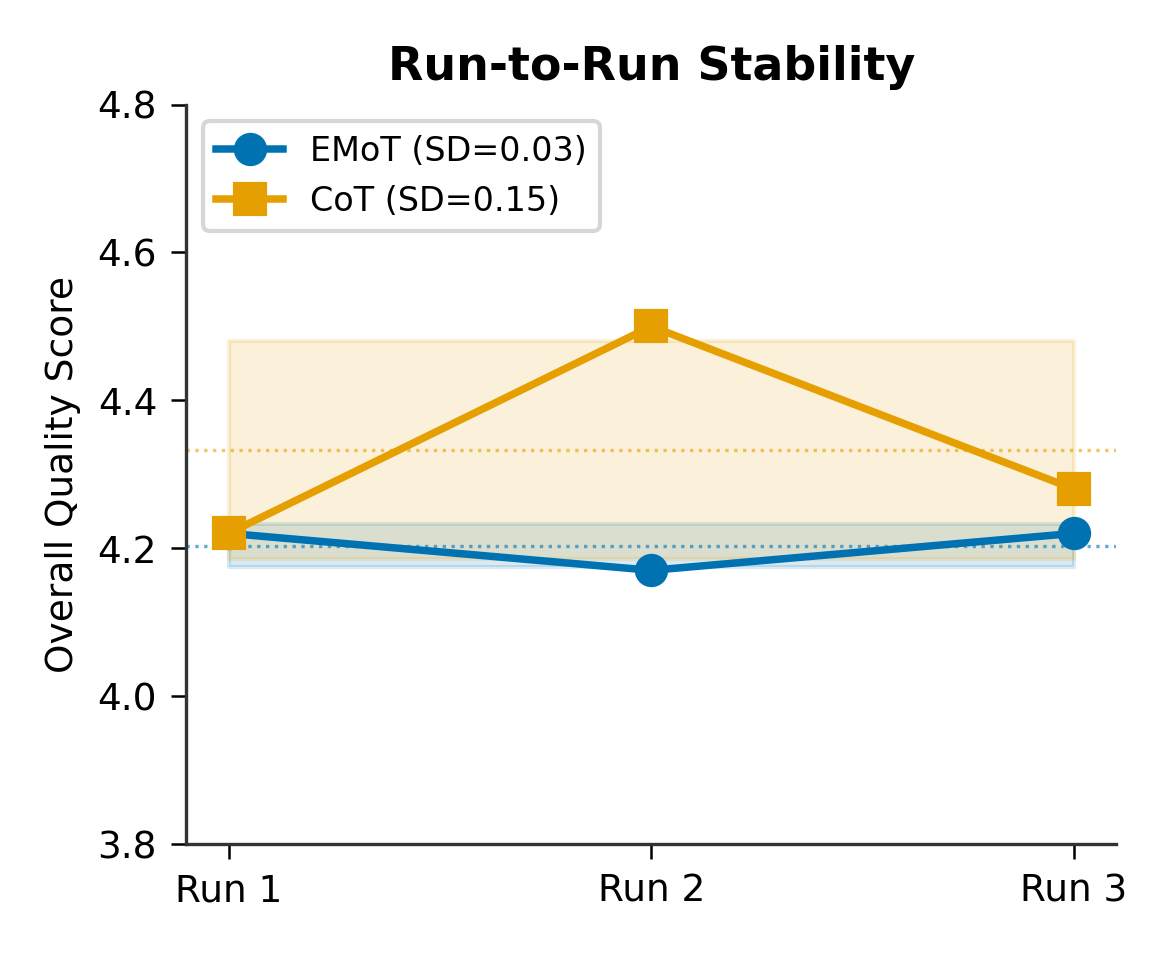}
\caption{Run-to-run stability comparison. EMoT produces identical overall scores across all three independent runs (SD=0.00), while CoT exhibits greater variability (SD=0.15).}
\label{fig:stability}
\end{figure}

Across three independent blind evaluations, CoT outperformed EMoT on overall quality by a margin of 0.13 points (4.33 vs.\ 4.20). Figure~\ref{fig:radar} illustrates the per-criterion comparison as a radar chart, showing the characteristic profile of each method. We note that these scores are derived from ordinal Likert scales (1--5) where arithmetic means should be interpreted cautiously. Two findings merit particular attention. First, EMoT demonstrated remarkable stability across runs, producing an identical overall score of 4.2 in all three evaluations (SD=0.00), while CoT exhibited greater variability (SD=0.15), as shown in Figure~\ref{fig:stability}. This suggests that EMoT's hierarchical architecture may produce more consistent reasoning outputs, though this observation requires confirmation with larger samples. Second, EMoT outperformed CoT on Cross-Domain Synthesis (4.8 vs.\ 4.4), the only criterion on which EMoT achieved a clear and consistent advantage. Both systems reached parity on Structured Output (5.0 vs.\ 5.0). EMoT's weakest criterion remained Dormant Thought Management (2.9), indicating that while the dormancy mechanism is architecturally present, its contribution is not yet fully reflected in the quality of final outputs as assessed by the judge.

\subsection{Ablation Studies}

To assess the contribution of EMoT's two principal architectural components, we conducted ablation experiments disabling each in isolation while holding all other parameters constant. Each ablation condition was evaluated once across the same three test cases used in the quality benchmark (Patient Bengt, Climate Migration, Vaccine Prioritisation), using the identical Claude Sonnet~4 judge setup described in Section~5.1.

\begin{table}[!ht]
\centering
\caption{Ablation study results: overall quality scores (1--5 scale).}
\label{tab:ablation}
\begin{tabular}{lcc}
\toprule
\textbf{Configuration} & \textbf{Overall Score} & \textbf{Delta from Full EMoT} \\
\midrule
Full EMoT & 4.20 & -- \\
No Dormancy (\texttt{-{}-no-dormancy}) & 1.00 & $-3.20$ \\
No Memory Palace (\texttt{-{}-no-memory-palace}) & 4.10 & $-0.10$ \\
\bottomrule
\end{tabular}
\end{table}

\textbf{Dormancy ablation.} Disabling the Strategic Dormancy Controller (\texttt{-{}-no-dormancy}) produced a catastrophic drop in quality from 4.20 to 1.00/5.0 across all evaluation criteria. The 1.0 score represents the floor of the judge rubric (minimum on all six criteria), indicating that the system produced no meaningful output rather than merely degraded output. For reference, standard CoT scored 4.33 under the same judge, illustrating that the no-dormancy configuration performs far below even a simple single-pass baseline. With dormancy disabled, the CEO module's pruning mechanism permanently eliminated all nodes falling below the Trust Score threshold, preventing any solution from being synthesised. This result demonstrates that strategic dormancy is not a decorative feature but an architecturally essential component: without the capacity to preserve and reactivate low-confidence nodes, EMoT's hierarchical processing collapses entirely. The SDC functions as a necessary counterbalance to the CEO's efficiency-driven pruning, ensuring that potentially valuable reasoning paths survive long enough to contribute to the final synthesis.

\textbf{Memory Palace ablation.} Disabling the Memory Palace (\texttt{-{}-no-memory-palace}) produced a modest but measurable decrease in overall quality from 4.20 to 4.10/5.0. The effect was most pronounced on Memory Utilisation and Cross-Domain Synthesis, the two criteria most directly served by persistent mnemonic encoding. This suggests that the Memory Palace contributes incremental value to EMoT's reasoning quality, primarily by enabling cross-iteration retrieval of insights that would otherwise be lost between processing cycles.

\begin{figure}[!ht]
\centering
\includegraphics[width=\columnwidth]{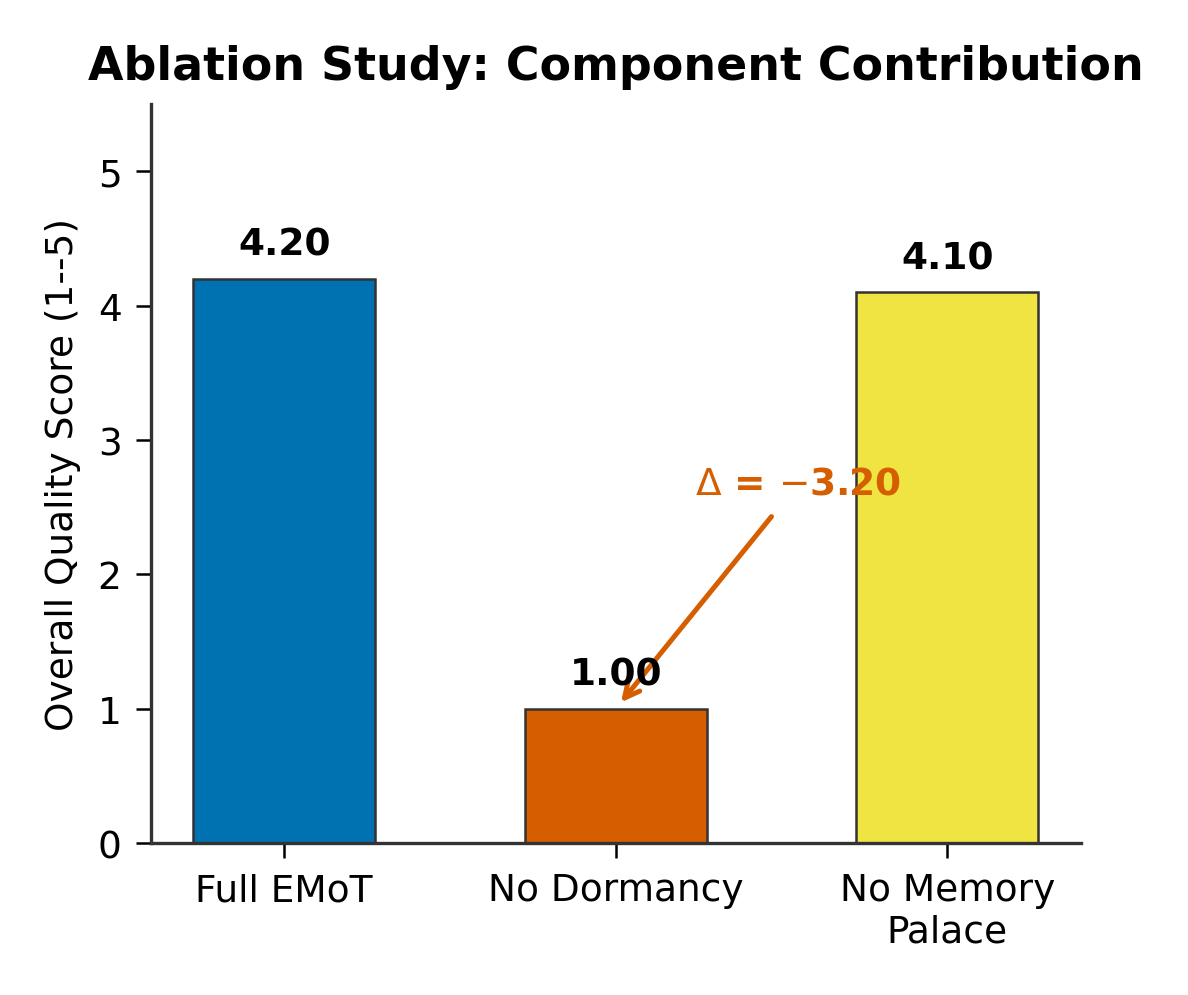}
\caption{Ablation study results. Disabling the Strategic Dormancy Controller causes a catastrophic quality drop from 4.20 to 1.00 ($-3.20$), while removing the Memory Palace produces a modest decrease to 4.10 ($-0.10$).}
\label{fig:ablation}
\end{figure}

Figure~\ref{fig:ablation} visualises the ablation results, highlighting the catastrophic $-3.20$ drop when dormancy is disabled. These ablation results provide direct evidence that dormancy is architecturally essential in this implementation: the system cannot produce meaningful output without the SDC counterbalancing the CEO's pruning. However, these results do not demonstrate the quality or frequency of dormancy-driven reactivation in practice; per-node reactivation statistics (e.g., how many dormant nodes were reactivated and what proportion of final-output insights they contributed) were not collected and remain an important direction for future instrumentation.

\subsection{Illustrative Clinical Case: Patient Bengt}

The clinical reasoning case provides an illustrative demonstration of EMoT's cross-domain integration capabilities. ``Patient Bengt'' is a 76-year-old male presenting with progressive multi-system deterioration over one year, including neurological symptoms (headache, dizziness with falls, forgetfulness), systemic symptoms (fatigue, weakness, poor appetite, nausea), and respiratory symptoms (cough). The patient has type~2 diabetes managed with metformin (HbA1c 52 mmol/mol, indicating reasonably well-controlled disease) and was previously on B12 and folate supplementation, which was discontinued due to supply chain disruption.

\begin{table}[!ht]
\centering
\caption{Per-case scores for Patient Bengt (blind evaluation, Run~1).}
\label{tab:bengt_scores}
\begin{tabular}{lccc}
\toprule
\textbf{Criterion} & \textbf{EMoT} & \textbf{CoT} & \textbf{Winner} \\
\midrule
Recursion Depth & 4 & 4 & Tie \\
Dormant Thought Management & 3 & 2 & EMoT \\
Cross-Domain Synthesis & 4 & 3 & EMoT \\
Memory Utilisation & 4 & 4 & Tie \\
Structured Output & 5 & 5 & Tie \\
Solution Quality & 4 & 4 & Tie \\
\midrule
\textbf{Average} & \textbf{4.00} & \textbf{3.67} & \textbf{EMoT} \\
\bottomrule
\end{tabular}
\end{table}

EMoT produced higher scores than CoT on this case with an average score of 4.00 vs.\ 3.67 in this preliminary evaluation. The advantages were in Dormant Thought Management (3 vs.\ 2) and Cross-Domain Synthesis (4 vs.\ 3). EMoT identified diabetes complications as an initial hypothesis, strategically set it aside given reasonably well-controlled glycaemic status, and maintained pernicious anaemia as a dormant possibility to revisit if the primary treatment response was poor. The cross-domain synthesis advantage reflected EMoT's capacity to integrate haematology, geriatrics, pharmacology, and supply chain perspectives into a unified diagnostic framework. CoT, by contrast, followed a linear diagnostic progression without explicitly managing alternative hypotheses.

Both systems correctly identified B12 deficiency as the primary diagnosis, connecting the temporal correlation between supplement discontinuation and symptom onset with the borderline-low B12 level (291~pmol/L; reference range: 150--750~pmol/L; clinically ambiguous zone: 150--300~pmol/L \cite{ref32}). Definitive diagnosis would require measurement of methylmalonic acid (MMA) and homocysteine levels to confirm functional B12 deficiency, as serum B12 alone has limited sensitivity in the borderline range. EMoT's output integrated clinical, pharmacological, and healthcare systems perspectives into a unified diagnostic framework, recognising B12 deficiency secondary to the combined effects of metformin-induced malabsorption and supplement supply disruption.

We emphasise that EMoT is not a clinical decision support system and has not been validated for clinical use. No clinical expert reviewed the diagnostic outputs for medical correctness or safety. The Patient Bengt case serves as an illustrative demonstration of cross-domain reasoning capabilities, not as clinical validation. Furthermore, while LLM-generated outputs may introduce novel terminology or framings, such neologisms require clinical validation before being considered established medical concepts.

\subsection{Cross-Model Sanity Check}

To provide a basic sanity check on whether EMoT's performance is attributable to the underlying LLM rather than the architecture, we ran the clinical reasoning case (Patient Bengt) through EMoT using Claude Sonnet~4 and a deterministic stub baseline. This is not a meaningful comparative benchmark but rather a verification that the architecture contributes beyond the stub's template-based responses.

\begin{table}[!ht]
\centering
\caption{Cross-model sanity check results.}
\label{tab:cross_model}
\begin{tabular}{lcccc}
\toprule
\textbf{Model} & \textbf{Tier} & \textbf{Avg Score} & \textbf{Runtime} & \textbf{Tokens} \\
\midrule
Claude Sonnet 4 & Advanced & 3.50 & 176 s & 19,438 \\
Stub (baseline) & Baseline & 1.00 & 0.1 s & 0 \\
\bottomrule
\end{tabular}
\end{table}

We note that the Claude Sonnet~4 score in this evaluation run (3.50) differs from the blind-judge benchmark score for the same case (4.00 in Table~\ref{tab:bengt_scores}). This divergence likely reflects differences in prompt templates and random seeds across evaluation runs, as these represent separate executions with different configurations. This variability further underscores the need for larger-scale evaluations with multiple runs per condition.

Claude Sonnet~4 produced a clinically correct diagnosis of ``Cobalamin (B12) deficiency with early megaloblastic anaemia, precipitated by medication supply disruption and compounded by metformin-induced B12 malabsorption.'' The system generated 73 insights across 13 active nodes in 33 LLM calls, with an internal quality score of 0.671.

The cross-domain synthesis score (4/5) suggested effective integration across geriatrics, endocrinology, haematology, and pharmacology. The structured output score (5/5) demonstrated that EMoT's hierarchical processing can produce well-organised clinical assessments. The lower scores on recursion depth (3/5) and dormant thought management (2/5) suggest room for improvement in EMoT's prompting strategy for single-case evaluations.

\subsection{Multi-Technique Accuracy Benchmark}

To complement the qualitative LLM-as-Judge evaluation, we conducted a multi-technique reasoning benchmark comparing EMoT against three baselines (Direct prompting, CoT, and Self-Consistency) on 15 short-answer problems drawn from five categories: mathematical reasoning (3 problems), logical reasoning (3), multi-step question answering (3), planning (3), and BIG-Bench Hard (BBH) tasks (3). We include this benchmark not because EMoT is intended for such tasks, but to explicitly characterise the failure mode and overthinking behaviour on simple problems. All evaluations used Claude Sonnet~4 with improved answer extraction to ensure that accuracy differences reflect genuine reasoning quality rather than parsing artefacts.

\begin{table}[!ht]
\centering
\caption{Multi-technique accuracy benchmark: correct answers per category (3 problems each, 15 total).}
\label{tab:accuracy_benchmark}
\small
\begin{tabularx}{\columnwidth}{@{}lXXXXXX@{}}
\toprule
\textbf{Technique} & \textbf{Math} & \textbf{Logic} & \textbf{M-QA} & \textbf{Plan.} & \textbf{BBH} & \textbf{Total} \\
\midrule
Direct & 3/3 & 3/3 & 3/3 & 3/3 & 3/3 & 15/15 (100\%) \\
CoT & 3/3 & 1/3 & 3/3 & 2/3 & 2/3 & 11/15 (73\%) \\
Self-Cons. & 3/3 & 1/3 & 3/3 & 1/3 & 1/3 & 9/15 (60\%) \\
EMoT & 1/3 & 1/3 & 1/3 & 1/3 & 0/3 & 4/15 (27\%) \\
\bottomrule
\end{tabularx}
\normalsize
\end{table}

\begin{table}[!ht]
\centering
\caption{Efficiency comparison across techniques (averaged over 15 problems).}
\label{tab:efficiency}
\begin{tabular}{lccc}
\toprule
\textbf{Technique} & \textbf{Avg Tokens} & \textbf{Avg Time} & \textbf{Cost (15 problems)} \\
\midrule
Direct & 81 & 1.8s & \$0.01 \\
CoT & 414 & 6.4s & \$0.06 \\
Self-Consistency & 1,236 & 26.9s & \$0.17 \\
EMoT & 12,136 & 183.4s & \$1.64 \\
\bottomrule
\end{tabular}
\end{table}

The accuracy results reveal a clear pattern: EMoT underperforms all baselines on short-answer reasoning tasks requiring precise factual or mathematical answers. Notably, the improved answer extraction used in this evaluation round actually produced slightly lower EMoT accuracy (27\%) compared to the earlier extraction method (40\%), confirming that the poor performance reflects genuine reasoning degradation rather than an extraction artefact. Analysis of EMoT's incorrect responses reveals a systematic overthinking effect: for simple arithmetic problems (e.g., ``2 bolts + 1 bolt = ?''), EMoT activates 13 specialised nodes that generate analyses of supply chain implications, quality control considerations, and economic factors, while the straightforward calculation is lost in the synthesis.

\begin{figure}[!ht]
\centering
\includegraphics[width=\columnwidth]{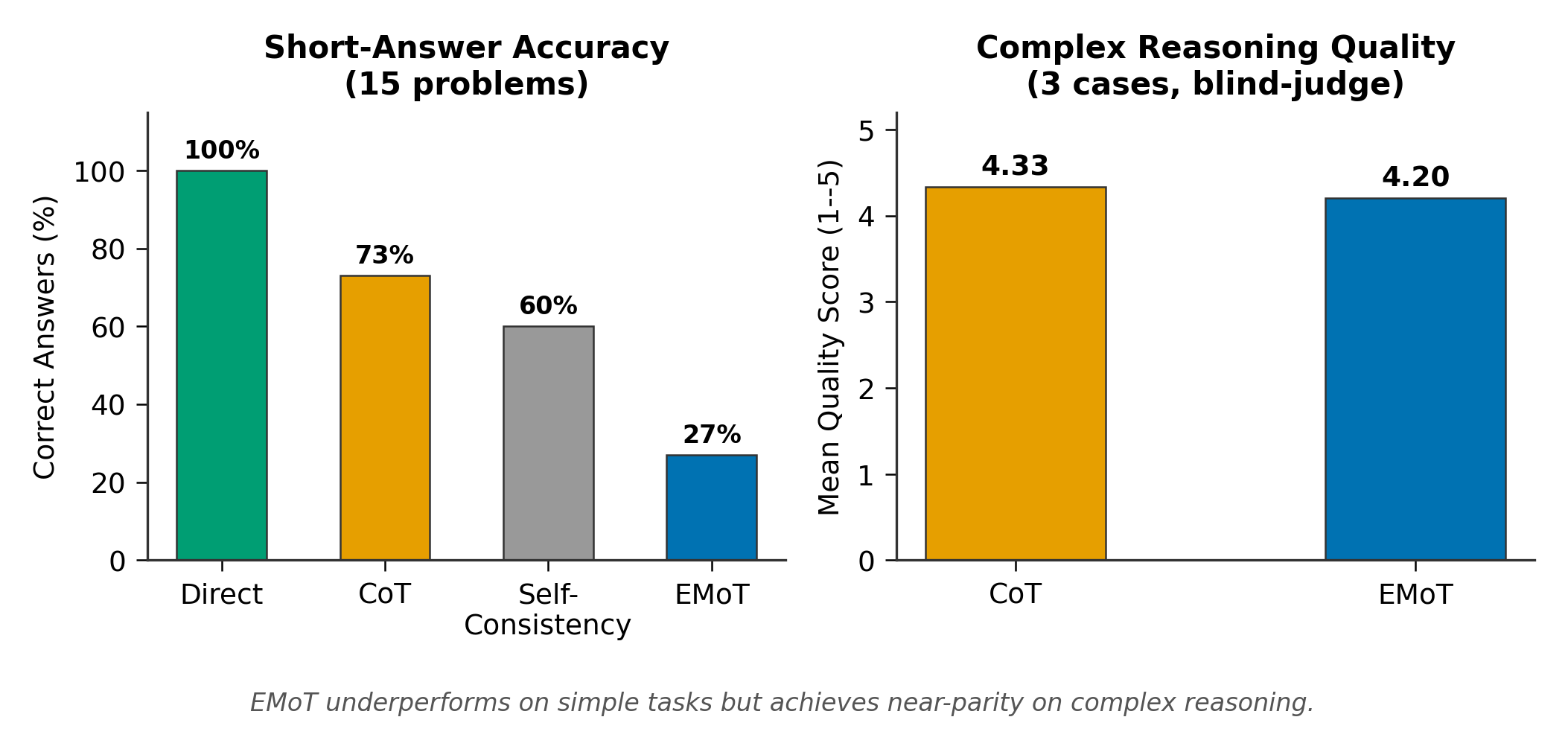}
\caption{Trade-off between short-answer accuracy (multi-technique benchmark) and complex reasoning quality (blind-judge benchmark). EMoT underperforms on factual accuracy but achieves near-parity with CoT on complex multi-domain reasoning quality.}
\label{fig:accuracy_quality}
\end{figure}

These results are consistent with our thesis that EMoT is not designed as a general-purpose reasoning enhancer but rather as a research prototype exploring reasoning infrastructure for complex, multi-domain problems. Figure~\ref{fig:accuracy_quality} illustrates this trade-off: on short-answer tasks, EMoT's accuracy falls well below simpler methods, yet on the complex reasoning quality benchmark (Section~5.2), EMoT achieves near-parity with CoT. On factual short-answer tasks, the architectural overhead introduces both cost and accuracy penalties.

Notably, the Direct prompting baseline achieves 100\% accuracy, outperforming both CoT (73\%) and Self-Consistency (60\%). This suggests that for a capable model such as Claude Sonnet~4, explicit step-by-step reasoning can introduce errors on simple problems through overthinking. This overthinking phenomenon has been increasingly documented: Chen et al.\ \cite{ref38} showed that o1-like models allocate disproportionate reasoning tokens to trivial problems, degrading both efficiency and accuracy; Sui et al.\ \cite{ref39} identified similar satisficing failures in search-augmented reasoning; and Hebenstreit et al.\ \cite{ref26} found that chain-of-thought strategies can degrade performance compared to direct prompting on certain task categories. EMoT's overthinking effect on simple tasks (activating 13 specialised nodes for ``2+3=?'') represents an extreme case of this phenomenon, driven by architectural rather than model-level over-reasoning.

\subsection{Computational Cost}

EMoT's hierarchical processing incurs substantially higher computational costs than single-pass CoT:

\begin{table}[!ht]
\centering
\caption{Computational cost comparison (3 test cases).}
\label{tab:cost}
\begin{tabular}{lccc}
\toprule
\textbf{Metric} & \textbf{EMoT} & \textbf{CoT} & \textbf{Ratio} \\
\midrule
LLM calls & 99 & 3 & 33x \\
Tokens (EMoT engine) & ${\sim}79{,}052$ & ${\sim}3{,}000$ (est.) & ${\sim}26$x \\
Runtime & ${\sim}1{,}214$ s & ${\sim}97$ s & ${\sim}13$x \\
Cost estimate & ${\sim}\$0.36$ & ${\sim}\$0.02$ (est.) & ${\sim}18$x \\
\bottomrule
\end{tabular}
\end{table}

EMoT required approximately 33 LLM calls per test case (99 total across 3 cases) compared to CoT's single call per case. The token overhead was approximately 26-fold, and wall-clock runtime was approximately 13-fold higher. These costs reflect EMoT's three-iteration processing cycle across its four-level hierarchy. Cost estimates are based on Claude Sonnet~4 API pricing at the time of evaluation.

% ============================================================
% 6. DISCUSSION
% ============================================================
\section{Discussion}

\subsection{When EMoT May Offer Advantages}

Our preliminary results suggest that EMoT's advantages may be most pronounced in problems exhibiting the following characteristics:

\begin{itemize}
    \item \textbf{Multi-domain problems requiring cross-disciplinary synthesis}, where insights from distinct knowledge areas must be integrated into a coherent solution.
    \item \textbf{High diagnostic uncertainty}, where initial hypotheses may be wrong and the reasoning system must revise its assessment as evidence accumulates.
    \item \textbf{Cases with hidden or overlooked factors} (such as supply chain effects on medication availability) that require cross-domain thinking to identify.
    \item \textbf{Problems where premature pruning carries high cost}, i.e., where discarding a low-probability hypothesis early may foreclose the correct solution.
\end{itemize}

The ablation studies (Section~5.3) provide direct evidence for the architectural necessity of strategic dormancy: disabling the SDC caused a catastrophic quality drop from 4.2 to 1.0, demonstrating that dormancy is not merely a theoretical differentiator but a load-bearing architectural component without which the system cannot function.

\begin{figure}[!ht]
\centering
\includegraphics[width=\columnwidth]{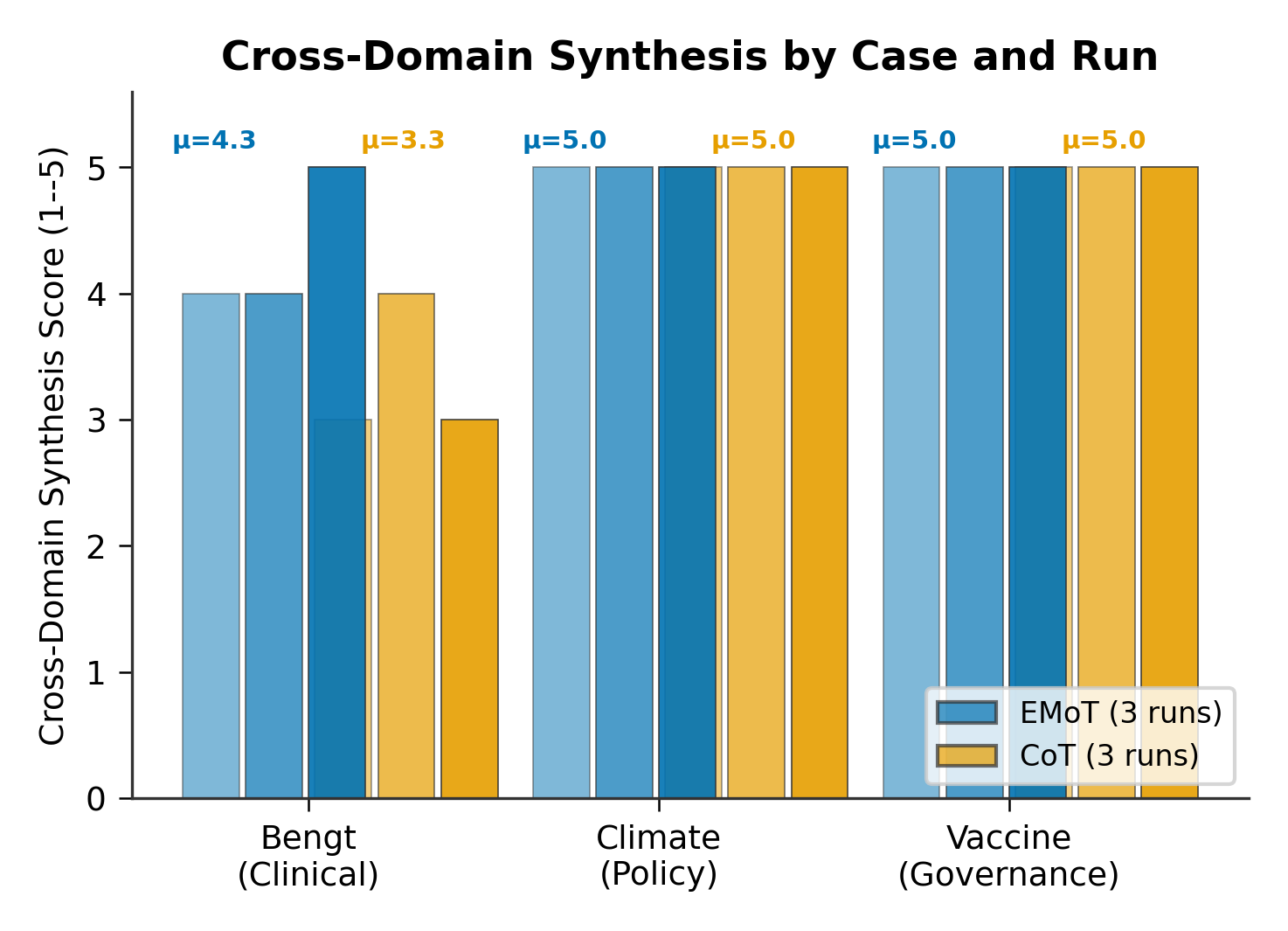}
\caption{Per-case, per-run Cross-Domain Synthesis scores. EMoT's advantage is most consistent on the clinical reasoning case (Patient Bengt), where cross-domain integration is most demanding.}
\label{fig:crossdomain}
\end{figure}

The clinical reasoning case, Patient Bengt, exemplifies all four characteristics. Figure~\ref{fig:crossdomain} shows the per-case, per-run Cross-Domain Synthesis scores, illustrating that EMoT's advantage is most consistent on the clinical case (mean 4.3 vs.\ 3.7), where cross-domain integration is most demanding. The correct diagnosis required integrating haematology (B12 deficiency), endocrinology (diabetes management, metformin effects), neurology (neuropsychiatric symptoms), and supply chain management (supplement availability). EMoT's hierarchical processing and dormancy management enabled it to maintain multiple diagnostic hypotheses simultaneously and to integrate clinical, pharmacological, and healthcare systems perspectives into a unified diagnostic framework.

This finding aligns with research on expert diagnostic reasoning, which emphasises the importance of maintaining differential diagnoses and reconsidering initially dismissed possibilities \cite{ref13}. EMoT's dormancy mechanism provides a computational analogue to the cognitive strategy Croskerry describes: in his dual-process model, diagnostic errors frequently arise when rapid Type~1 processing (pattern recognition, heuristic-driven) prematurely closes the hypothesis space before slower, deliberate Type~2 processing (analytical, evidence-weighing) can reconsider alternatives. EMoT's dormancy-reactivation cycle maps to this Type~1-to-Type~2 transition: initial fast assessment may deprioritise a hypothesis (dormancy), but the system retains the capacity to reactivate it when deeper analysis reveals inconsistencies, much as an experienced clinician revisits a dismissed diagnosis when the clinical picture evolves unexpectedly. The ablation evidence strengthens this argument considerably: when dormancy is removed, the system does not merely degrade but collapses entirely (1.0/5.0), because the CEO's pruning mechanism eliminates all nodes without the SDC's counterbalancing preservation. This interdependence between efficiency-driven pruning and dormancy-driven preservation may represent a general design principle for reasoning architectures.

A fourth case, Patient Erik (Appendix~A.4), was designed as a ``diagnostic trap'' to test this thesis more directly. Erik presents with worsening atrial fibrillation and heart failure; the GP's intuitive response -- doubling the amiodarone dose -- is catastrophically wrong because amiodarone itself is causing thyrotoxicosis (AIT Type~2) through triple iodine loading (amiodarone, CT contrast from 4 months prior, undisclosed kelp supplements). The correct diagnosis requires integrating endocrinology, pharmacology, cardiology, radiology, and nutrition, and recognising that the treatment was the cause. Both EMoT and CoT correctly identified AIT Type~2, the triple iodine sources, and the iatrogenic harm (EMoT 4.17/5.0 vs.\ CoT 4.33/5.0). That both architectures solved this case reflects the strength of the underlying LLM (Claude Sonnet~4) rather than a limitation of the benchmark design; it suggests that for future evaluations, cases requiring temporal information arrival -- where hypotheses must genuinely be revised as new evidence emerges over multiple reasoning iterations -- may better isolate EMoT's dormancy advantage.

This finding also aligns with Jeon and Kim \cite{ref25}, who demonstrated that complex chain-of-thought prompting techniques do not significantly enhance performance over simpler approaches in clinical question answering. EMoT's advantage in clinical reasoning appears to derive not from more sophisticated prompting per se, but from its architectural capacity to maintain and cross-reference multiple diagnostic hypotheses simultaneously through the Memory Palace structure.

\subsection{When CoT Is Sufficient}

For two of the three test cases (policy design and AI governance), CoT produced higher scores than EMoT. These problems, while complex, have a more predictable structure: they require comprehensive coverage of known stakeholder perspectives and systematic analysis of policy options. CoT's linear, well-structured approach produced outputs that scored higher on recursion depth, cross-domain synthesis, and solution quality.

This suggests that EMoT's overhead may not be justified for problems where the reasoning path is relatively predictable and where the risk of prematurely discarding relevant hypotheses is low. The additional computational cost (33x LLM calls, 13x runtime) makes CoT the more efficient choice for such tasks.

\subsection{The Role of Memory Palace and Mnemonic Encoding}

The ablation study (Section~5.3) provides direct evidence for the Memory Palace's contribution: disabling it reduced overall quality from 4.20 to 4.1, with the effect concentrated on Memory Utilisation and Cross-Domain Synthesis. While this contribution is modest compared to the dormancy mechanism's essential role, it confirms that persistent mnemonic encoding adds measurable value. The five mnemonic encoding styles provide diverse retrieval pathways, though the current ablation does not isolate the contribution of individual encoding styles. Fine-grained ablation studies comparing performance with and without specific encoding styles represent an important direction for future work.

\subsection{Cost-Benefit Considerations}

EMoT's 33x LLM call overhead relative to CoT raises a legitimate question: under what circumstances would a practitioner accept this cost? We position EMoT explicitly as a research prototype exploring the architectural space of reasoning frameworks, not as a production tool for routine use. Similar to how many experimental architectures explore design spaces that inform future, more practical systems, EMoT's primary contribution lies in demonstrating that dormancy, mnemonic persistence, and hierarchical reasoning are viable architectural primitives for LLM-based reasoning. The insights gained from EMoT's architecture -- that preserving rather than pruning uncertain hypotheses may improve cross-domain reasoning, that persistent memory across iterations can surface overlooked connections -- are intended to inform future, more efficient implementations rather than to serve as a deployable system in their current form.

\subsection{Limitations}

Several limitations constrain the interpretation of our results:

\textbf{Evaluation methodology.} The LLM-as-Judge paradigm, while increasingly standard, introduces potential biases. Although we mitigated evaluation bias by removing method labels and conducting three independent runs per condition (providing variance estimates), the judge model remains from the same model family as the reasoning model, creating a potential self-preference bias \cite{ref15}. Wataoka et al.\ \cite{ref20} identified significant self-preference bias in LLM-as-Judge evaluations, where LLMs assign higher scores to outputs with lower perplexity, regardless of whether the outputs were self-generated. This self-preference effect may inflate all scores and may differentially affect EMoT and CoT scores in ways that are difficult to predict. Future evaluations should employ independent judge models from different model families (e.g., using GPT-4 or Gemini as judges for Claude-generated outputs) and, critically, include human expert assessment to complement the automated evaluation.

\textbf{Rubric circularity.} As noted in Section~5.1, two of the six evaluation criteria -- Dormant Thought Management and Memory Utilisation -- describe mechanisms that are architectural features of EMoT but not of CoT. This creates a potential circularity where EMoT is evaluated partly on criteria tailored to its own design. While CoT can score well on these criteria through implicit hypothesis management, the rubric may inherently favour architectures that make these processes explicit. Future evaluations should employ architecture-neutral criteria to enable fairer comparisons.

\textbf{Benchmark scope.} The quality evaluation covers three test cases across three domains with three independent runs per condition (providing variance estimates), and the multi-technique benchmark covers 15 short-answer problems across five categories. While the multiple runs strengthen confidence in the quality benchmark findings, both evaluations remain small-scale. Larger-scale evaluation on established benchmarks (GSM8K, MMLU, PubMedQA) with hundreds of items is needed to characterise EMoT's strengths and weaknesses more precisely.

\textbf{Computational cost.} EMoT's 33x LLM call overhead and 13x runtime increase represent a significant practical limitation. While the CEO module targets efficiency gains, the current implementation remains substantially more expensive than CoT for comparable tasks.

\textbf{Single-model evaluation.} Our primary benchmarks used a single LLM (Claude Sonnet~4). Performance characteristics may differ with other models, particularly smaller or open-source models deployed locally.

\textbf{Dormancy utilisation.} The ablation study (Section~5.3) provides strong evidence that dormancy is architecturally essential, but the empirical evidence for the quality of dormancy-driven reactivation in the final output remains limited to the Dormant Thought Management scores in our rubric (2.9/5.0). Direct measurement of reactivation frequency, the conditions triggering reactivation, and the specific impact of reactivated nodes on solution quality remains for future work.

\textbf{Clinical validity.} The clinical case (Patient Bengt) serves as an illustrative demonstration. EMoT is not a clinical decision support system and has not been validated by clinicians. No clinical expert reviewed the diagnostic outputs for medical correctness or safety. The diagnostic reasoning presented should not be interpreted as clinical advice or as evidence of clinical utility.

\subsection{Strategic Positioning}

EMoT is best understood not as a prompting technique but as a reasoning infrastructure -- an intermediate layer between the LLM and the problem that provides persistent memory, hierarchical organisation, and strategic thought management. In this sense, EMoT is analogous to an operating system for structured reasoning: it manages cognitive resources, schedules processing across hierarchical levels, and maintains state across reasoning episodes.

This positioning suggests that EMoT and simpler prompting techniques are not mutually exclusive. EMoT could incorporate CoT as the reasoning strategy within individual nodes, ToT as the expansion strategy within a single hierarchical level, or Self-Consistency as the evaluation strategy for competing Macro-level solutions.

% ============================================================
% 7. FUTURE WORK
% ============================================================
\section{Future Work}

Several directions for future development emerge from this work:

\textbf{Expanded empirical evaluation.} The multi-technique accuracy benchmark (Section~5.6) provides initial results on short-answer reasoning tasks, but larger-scale evaluation on established benchmarks (GSM8K for mathematical reasoning, MMLU for general knowledge, PubMedQA for biomedical question answering) with larger sample sizes is needed to characterise EMoT's performance boundaries more precisely. In particular, tasks requiring extended cross-domain synthesis, rather than factual short answers, may reveal contexts where EMoT's architectural overhead translates into measurable quality advantages.

\textbf{Dormancy and reactivation studies.} Designing evaluation scenarios that specifically require hypothesis reconsideration would test EMoT's dormancy mechanism more rigorously. Longitudinal problem-solving tasks where new information arrives over time represent a particularly promising test domain.

\textbf{On-premise deployment.} Deploying EMoT with local open-source models (Qwen-3 7B/14B on GB10 nodes with RTX 4090-class GPUs) would enable privacy-preserving reasoning for sensitive clinical and research applications. Initial integration with the Ollama backend is implemented but has not yet been benchmarked.

\textbf{Knowledge graph integration.} Connecting EMoT's Memory Palace with external knowledge graphs (e.g., biomedical ontologies, clinical guidelines) could enhance the quality and verifiability of reasoning, particularly in specialised domains.

\textbf{Multi-agent orchestration.} Deploying multiple EMoT instances as specialised agents (e.g., a clinical reasoning agent, a regulatory compliance agent, a systems-thinking agent) coordinated through a meta-level orchestrator could enable more complex problem-solving than a single EMoT instance.

\textbf{Integration with orchestration frameworks.} EMoT's current monolithic implementation limits scalability, observability, and composability. Reimplementing EMoT's architecture on top of graph-based agent orchestration frameworks such as LangGraph \cite{ref27} could address several of these limitations. LangGraph's persistent typed state, conditional edges, parallel branch execution, human-in-the-loop interrupts, and built-in tracing infrastructure map naturally to EMoT's requirements for state management, dormancy logic, parallelised node processing, clinical oversight, and runtime observability. We hypothesise that such an implementation would retain EMoT's architectural advantages while achieving significant improvements in wall-clock time through parallelisation and enabling deployment patterns that the current monolithic implementation cannot support.

\textbf{Efficiency optimisation.} The CEO module's pathway pruning and result caching mechanisms require empirical tuning to reduce the current 33x LLM call overhead while preserving reasoning quality.

\textbf{Testable hypotheses.} Two concrete hypotheses emerge from this work that can guide future evaluation design. First, on tasks requiring sequential information arrival and hypothesis revision, we hypothesise that EMoT will outperform CoT on judge-rated solution quality at comparable token budgets. Second, on GSM8K-style short problems, we hypothesise that EMoT will remain inferior to direct prompting, reinforcing its non-general-purpose positioning. Confirmation or refutation of these hypotheses would substantially clarify the boundary conditions for EMoT's utility.

% ============================================================
% 8. CONCLUSION
% ============================================================
\section{Conclusion}

We have presented the Enhanced Mycelium of Thought (EMoT) framework, a bio-inspired reasoning architecture that introduces hierarchical network organisation, strategic dormancy with reactivation, and mnemonic memory encoding to LLM-based reasoning. Our preliminary evaluation (Sections~5.2--5.6) reveals a characteristic trade-off: EMoT achieves near-parity with CoT on complex, multi-domain problems (with a consistent advantage on Cross-Domain Synthesis), but substantially underperforms simpler methods on short-answer tasks and incurs significant computational overhead. The ablation studies (Section~5.3) identify strategic dormancy as architecturally essential, while the accuracy benchmark (Section~5.6) confirms that EMoT's overhead is counterproductive for problems that do not require hypothesis revision or cross-domain integration. These findings should be interpreted with caution given the limitations detailed in Sections~5.1 and~6.5, including small sample sizes, LLM-as-Judge evaluation, and substantial computational overhead.

EMoT's primary contribution is architectural rather than empirical: it demonstrates that dormancy, mnemonic persistence, and hierarchical reasoning are viable primitives for LLM-based reasoning infrastructure. As such systems are increasingly applied to high-stakes domains, architectures that preserve rather than prune uncertain hypotheses and maintain persistent memory across reasoning iterations may prove valuable, though extensive validation will be required before any such system could be considered for deployment in clinical or policy settings.

The complete source code, benchmark suite, and evaluation data will be made available upon acceptance.

% ============================================================
% DECLARATIONS
% ============================================================
\section*{Declarations}

\textbf{Conflict of Interest:} The author declares no competing interests.

\textbf{Funding:} This research received no external funding.

\textbf{Data Availability:} The source code, benchmark suite, and evaluation data will be made available in a public repository upon publication.

\textbf{Ethics Statement:} This study involved no human participants, patient data, or animal subjects. The clinical case (Patient Bengt) is a synthetic construct designed for evaluation purposes.

% ============================================================
% REFERENCES
% ============================================================

% ============================================================
% APPENDIX A
% ============================================================
\appendix
\section{Benchmark Case Descriptions}

\subsection{Case 1: Clinical Reasoning (Patient Bengt)}

A 76-year-old male presenting with progressive multi-system deterioration over 12 months. Symptoms include headache, dizziness with falls, increased forgetfulness, fatigue, weakness, poor appetite, nausea, and cough. Medical history includes type~2 diabetes managed with metformin (HbA1c 52 mmol/mol, approximately 6.9\% DCCT, indicating reasonably well-controlled disease). Previously prescribed B12 and folate supplementation, discontinued approximately 12 months ago due to pharmacy supply chain disruption. Laboratory findings: B12 291~pmol/L (reference range: 150--750~pmol/L; clinically ambiguous zone: 150--300~pmol/L \cite{ref32}), folate 8~nmol/L. The task requires identifying the most likely diagnosis, recommending essential investigations (including methylmalonic acid and homocysteine to confirm functional B12 deficiency), and proposing a management plan.

\subsection{Case 2: Policy Design (Climate Migration)}

Design a comprehensive global strategy addressing projected climate-induced displacement of 200+ million people by 2050 \cite{ref34}. The strategy must address: (a) international legal frameworks, (b) economic mechanisms for managed relocation, (c) integration of displaced populations, (d) political feasibility across different governance systems, (e) environmental monitoring and early warning, and (f) cultural preservation during displacement.

\subsection{Case 3: AI Governance (Pandemic Vaccine Prioritisation)}

Design an evidence-based AI policy model for vaccine prioritisation and distribution during a pandemic. The model must balance: (a) epidemiological optimisation, (b) equity across demographic groups, (c) supply chain constraints, (d) public trust and acceptance, (e) real-time adaptation to emerging variants, and (f) international coordination.

\subsection{Case 4: Diagnostic Complexity (Patient Erik)}

A 71-year-old retired shipyard electrician presenting with worsening exertional dyspnoea, palpitations, weight loss (4~kg over 6 weeks), tremor, heat intolerance, and emotional lability. Timeline: diagnosed with paroxysmal atrial fibrillation 18 months ago, started on amiodarone 200~mg daily with good rhythm control; CT pulmonary angiography with iodinated contrast 4 months ago (PE excluded); wife started kelp-based ``thyroid support'' supplements (${\sim}500$~mcg iodine/day, undisclosed) 3 months ago; GP interpreted AF recurrence 2 weeks ago as amiodarone failure, doubled the dose to 400~mg and added digoxin 125~mcg. On examination: HR 138, BP 154/68 (wide pulse pressure), fine tremor, lid lag, mild proptosis, warm moist skin, hyperreflexia, signs of heart failure (raised JVP, bilateral oedema, bibasal crackles). Laboratory: TSH $<0.01$~mU/L (suppressed), fT4 58~pmol/L (ref 12--22), fT3 18.2~pmol/L (ref 3.1--6.8), BNP 890~pg/mL, IL-6 28~pg/mL, thyroid antibodies negative. Thyroid ultrasound: decreased Doppler vascularity.

This case was designed as a ``diagnostic trap'' to test multi-domain reasoning. The correct diagnosis is amiodarone-induced thyrotoxicosis (AIT) Type~2, precipitated by triple iodine loading from three sources (amiodarone, CT contrast, kelp supplements), causing thyrotoxic cardiomyopathy. The GP's dose escalation represents iatrogenic harm. Distinguishing AIT Type~1 (increased vascularity, thionamides) from Type~2 (decreased vascularity, corticosteroids) requires integrating endocrinology, pharmacology, cardiology, radiology, and nutrition. The case tests whether reasoning architectures can identify that the treatment itself was the cause of deterioration.

% ============================================================
% APPENDIX B
% ============================================================
\section{Full Scoring Tables}

Note: Each run evaluates 3 cases (Patient Bengt, Climate Migration, Vaccine Prioritisation). Per-run overall scores in Table~\ref{tab:per_run_scores} are computed as the mean of per-case averages. Per-criterion means in Table~\ref{tab:per_criterion_means} are computed across all 9 case-evaluations (3 runs $\times$ 3 cases).

\subsection{Per-Case Scores for All 3 Blind-Judge Runs}

\textbf{Run 1} (Overall: EMoT 4.22 vs.\ CoT 4.22 -- Tie)

\begin{table}[H]
\centering
\small
\begin{tabular}{llcc}
\toprule
\textbf{Case} & \textbf{Criterion} & \textbf{EMoT} & \textbf{CoT} \\
\midrule
Bengt & Recursion Depth & 4 & 4 \\
Bengt & Dormant Thought & 3 & 2 \\
Bengt & Cross-Domain Synthesis & 4 & 3 \\
Bengt & Memory Utilisation & 4 & 4 \\
Bengt & Structured Output & 5 & 5 \\
Bengt & Solution Quality & 4 & 4 \\
\textbf{Bengt} & \textbf{Average} & \textbf{4.00} & \textbf{3.67} \\
\midrule
Climate & Recursion Depth & 4 & 5 \\
Climate & Dormant Thought & 3 & 3 \\
Climate & Cross-Domain Synthesis & 5 & 5 \\
Climate & Memory Utilisation & 4 & 4 \\
Climate & Structured Output & 5 & 5 \\
Climate & Solution Quality & 5 & 5 \\
\textbf{Climate} & \textbf{Average} & \textbf{4.33} & \textbf{4.50} \\
\midrule
Vaccine & Recursion Depth & 4 & 5 \\
Vaccine & Dormant Thought & 3 & 3 \\
Vaccine & Cross-Domain Synthesis & 5 & 5 \\
Vaccine & Memory Utilisation & 4 & 4 \\
Vaccine & Structured Output & 5 & 5 \\
Vaccine & Solution Quality & 5 & 5 \\
\textbf{Vaccine} & \textbf{Average} & \textbf{4.33} & \textbf{4.50} \\
\bottomrule
\end{tabular}
\end{table}

\textbf{Run 2} (Overall: EMoT 4.17 vs.\ CoT 4.50 -- CoT)

\begin{table}[H]
\centering
\small
\begin{tabular}{llcc}
\toprule
\textbf{Case} & \textbf{Criterion} & \textbf{EMoT} & \textbf{CoT} \\
\midrule
Bengt & Recursion Depth & 4 & 4 \\
Bengt & Dormant Thought & 3 & 3 \\
Bengt & Cross-Domain Synthesis & 4 & 4 \\
Bengt & Memory Utilisation & 4 & 4 \\
Bengt & Structured Output & 5 & 5 \\
Bengt & Solution Quality & 4 & 4 \\
\textbf{Bengt} & \textbf{Average} & \textbf{4.00} & \textbf{4.00} \\
\midrule
Climate & Recursion Depth & 4 & 5 \\
Climate & Dormant Thought & 3 & 4 \\
Climate & Cross-Domain Synthesis & 5 & 5 \\
Climate & Memory Utilisation & 4 & 4 \\
Climate & Structured Output & 5 & 5 \\
Climate & Solution Quality & 4 & 5 \\
\textbf{Climate} & \textbf{Average} & \textbf{4.17} & \textbf{4.67} \\
\midrule
Vaccine & Recursion Depth & 4 & 5 \\
Vaccine & Dormant Thought & 3 & 4 \\
Vaccine & Cross-Domain Synthesis & 5 & 5 \\
Vaccine & Memory Utilisation & 4 & 5 \\
Vaccine & Structured Output & 5 & 5 \\
Vaccine & Solution Quality & 5 & 5 \\
\textbf{Vaccine} & \textbf{Average} & \textbf{4.33} & \textbf{4.83} \\
\bottomrule
\end{tabular}
\end{table}

\textbf{Run 3} (Overall: EMoT 4.22 vs.\ CoT 4.28 -- CoT)

\begin{table}[H]
\centering
\small
\begin{tabular}{llcc}
\toprule
\textbf{Case} & \textbf{Criterion} & \textbf{EMoT} & \textbf{CoT} \\
\midrule
Bengt & Recursion Depth & 4 & 4 \\
Bengt & Dormant Thought & 3 & 2 \\
Bengt & Cross-Domain Synthesis & 5 & 3 \\
Bengt & Memory Utilisation & 4 & 4 \\
Bengt & Structured Output & 5 & 5 \\
Bengt & Solution Quality & 5 & 4 \\
\textbf{Bengt} & \textbf{Average} & \textbf{4.33} & \textbf{3.67} \\
\midrule
Climate & Recursion Depth & 4 & 5 \\
Climate & Dormant Thought & 2 & 4 \\
Climate & Cross-Domain Synthesis & 5 & 5 \\
Climate & Memory Utilisation & 4 & 5 \\
Climate & Structured Output & 5 & 5 \\
Climate & Solution Quality & 4 & 5 \\
\textbf{Climate} & \textbf{Average} & \textbf{4.00} & \textbf{4.83} \\
\midrule
Vaccine & Recursion Depth & 4 & 4 \\
Vaccine & Dormant Thought & 3 & 3 \\
Vaccine & Cross-Domain Synthesis & 5 & 5 \\
Vaccine & Memory Utilisation & 4 & 4 \\
Vaccine & Structured Output & 5 & 5 \\
Vaccine & Solution Quality & 5 & 5 \\
\textbf{Vaccine} & \textbf{Average} & \textbf{4.33} & \textbf{4.33} \\
\bottomrule
\end{tabular}
\end{table}

\subsection{Per-Criterion Means Across 9 Case-Evaluations (3 Runs $\times$ 3 Cases)}

\begin{table}[H]
\centering
\begin{tabular}{lccc}
\toprule
\textbf{Criterion} & \textbf{EMoT Mean} & \textbf{CoT Mean} & \textbf{Winner} \\
\midrule
Recursion Depth & 4.0 & 4.6 & CoT \\
Dormant Thought Management & 2.9 & 3.1 & CoT \\
Cross-Domain Synthesis & 4.8 & 4.4 & EMoT \\
Memory Utilisation & 4.0 & 4.2 & CoT \\
Structured Output & 5.0 & 5.0 & Tie \\
Solution Quality & 4.6 & 4.7 & CoT \\
\midrule
\textbf{Overall} & \textbf{4.20 (SD=0.00)} & \textbf{4.33 (SD=0.15)} & \textbf{CoT} \\
\bottomrule
\end{tabular}
\end{table}

\subsection{Ablation Study Scores}

\begin{table}[H]
\centering
\begin{tabular}{lc}
\toprule
\textbf{Configuration} & \textbf{Overall Score} \\
\midrule
Full EMoT & 4.20 \\
No Dormancy (\texttt{-{}-no-dormancy}) & 1.00 \\
No Memory Palace (\texttt{-{}-no-memory-palace}) & 4.10 \\
\bottomrule
\end{tabular}
\end{table}

\subsection{Patient Erik (Diagnostic Complexity Case)}

\begin{table}[H]
\centering
\begin{tabular}{lccc}
\toprule
\textbf{Criterion} & \textbf{EMoT} & \textbf{CoT} & \textbf{Winner} \\
\midrule
Recursion Depth & 4 & 4 & Tie \\
Dormant Thought Management & 3 & 3 & Tie \\
Cross-Domain Synthesis & 5 & 5 & Tie \\
Memory Utilisation & 4 & 4 & Tie \\
Structured Output & 5 & 5 & Tie \\
Solution Quality & 4 & 5 & CoT \\
\midrule
\textbf{Average} & \textbf{4.17} & \textbf{4.33} & \textbf{CoT} \\
\bottomrule
\end{tabular}
\end{table}

Both EMoT and CoT correctly identified AIT Type~2 with triple iodine loading and iatrogenic harm from amiodarone dose escalation.

\begin{figure}[!ht]
\centering
\includegraphics[width=\columnwidth]{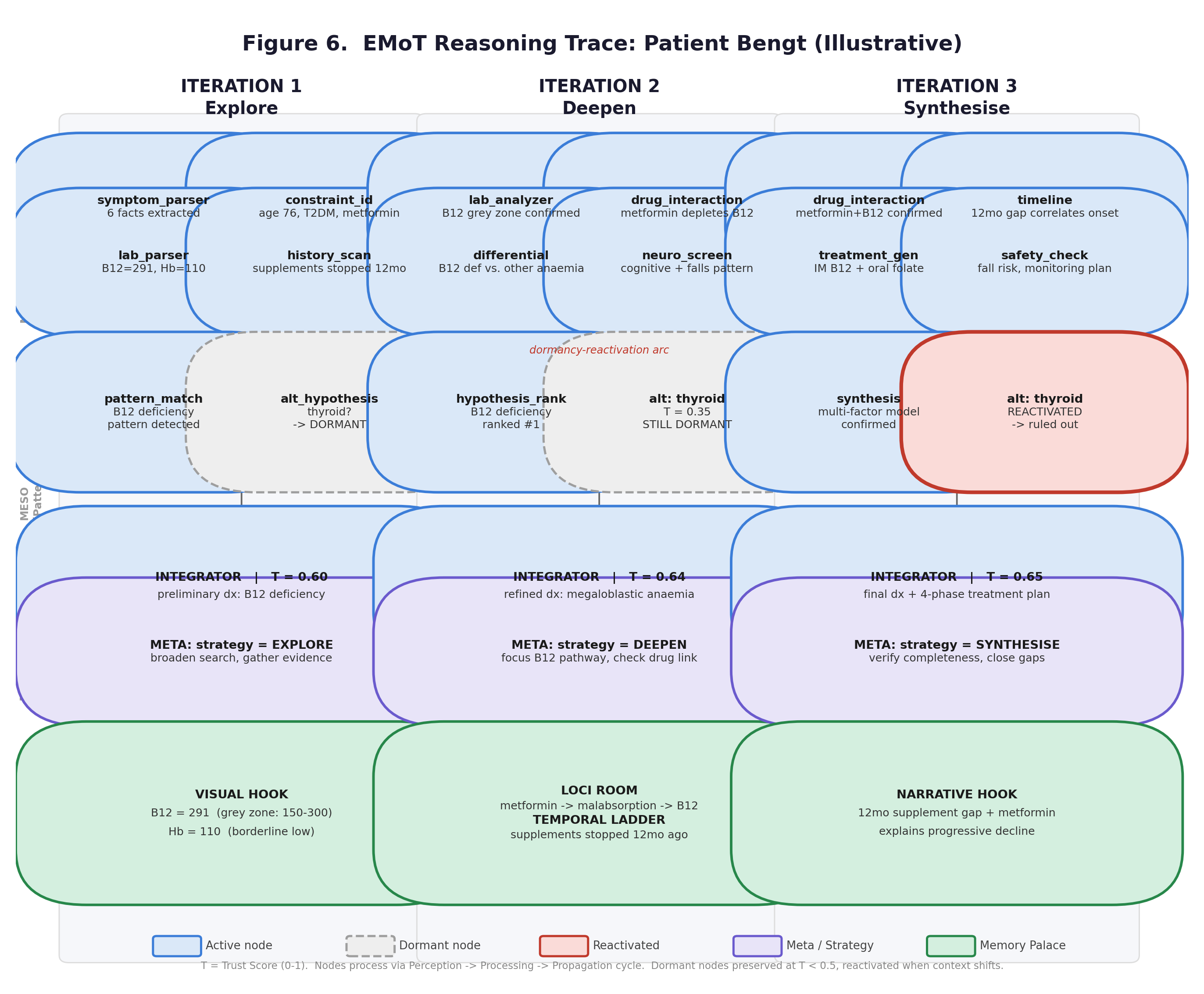}
\caption{Representative reasoning trace for the Patient Bengt case, illustrating EMoT's hierarchical processing across four levels. Dormant nodes (dashed borders) are preserved across iterations and reactivated when new evidence warrants reconsideration, demonstrating the Strategic Dormancy Controller's role in maintaining alternative hypotheses.}
\label{fig:reasoning_trace}
\end{figure}

\end{document}